# Employee Attrition Prediction

Rahul Yedida
PESIT-BSC, Bangalore
yrahul3910@gmail.com

Rahul Reddy
PESIT-BSC, Bangalore
rahul.r.nandyala@gmail.com

Rakshit Vahi
PESIT-BSC, Bangalore
vahi.rakshit@gmail.com

Rahul J
PESIT-BSC, Bangalore
janarahul123@gmail.com

Abhilash
PESIT-BSC, Bangalore
abhilash.ab04@gmail.com

Deepti Kulkarni
PESIT-BSC, Bangalore
deeptikulkarni211@gmail.com

*Abstract*—This project aims to predict whether an employee of a company will leave or not, using the k-Nearest Neighbors algorithm. We use evaluation of employee performance, average monthly hours at work and number of years spent in the company, among others, as our features. Other approaches to this problem include the use of ANNs, decision trees and logistic regression. The dataset was split, using 70% for training the algorithm and 30% for testing it, achieving an accuracy of 94.32%.

## I. Introduction

Employee resignations are a reality for any business. However, if the situation isn't handled properly, key staff members' departures can lead to a downturn in productivity. The organization may have to employ new people and train them on the tool that is being used, which is time consuming. Most organizations are interested in knowing which of their employees are at the risk of leaving.

This paper discusses the application of the k-Nearest Neighbours (KNN) algorithm as a method of predicting employee attrition. This is done by using data from Kaggle and treating the problem as a classification task. The conclusion is reached by comparing the performance of the KNN classifier against other techniques.

This paper is structured as follows. Section II discusses the attrition problem, and lists the work done by others using machine learning algorithms to solve the problem. Section III explores 4 different machine learning algorithms, including KNN, that this paper compares. Section IV outlines the experimental method employed in terms of the features used, pre-processing, and the metrics used to compare the algorithms. Section V presents the results of the comparison and a discussion of the same, and possible future work. Section VI concludes the paper by recommending the KNN classifier as an approach to solving the employee attrition prediction problem.

## II. Literature Survey

Employee attrition refers to the gradual loss of employees over time. Most literature on employee attrition categorizes it as either voluntary or involuntary. Involuntary attrition is thought of as the mistake of the employee, and refers to the organization firing the employee for various reasons. Voluntary attrition is when the employee leaves the organization by his own will.

This paper focuses on voluntary attrition. A meta-analytic review of voluntary attrition [1] found that the strongest predictors of voluntary attrition included age, pay, and job satisfaction. Other studies showed that several other features, such as working conditions, job satisfaction, and growth potential also contributed to voluntary attrition [2][3].

Organizations try to prevent employee attrition by using machine learning algorithms to predict the risk of an employee leaving, and then take pro-active steps for preventing such an incident.

## III. Methods

This paper discusses supervised learning methods of classification, since we know of the existence of two classes—working and left. This section outlines the theory behind each machine learning algorithm.

### A. Naive Bayes

Naive Bayes is a classification technique that has gained popularity due to its simplicity [4]. The Naive Bayes algorithm makes use of the assumption that all the variables are independent of each other, and then calculates probabilities, that are used for classification.

The algorithm works as follows: to get an output function Y given a set of input variables X, the algorithm estimates the values of $P(X|Y)$ and $P(Y)$, and then uses Bayes' rule to compute $P(Y|X)$, which is the required output, for each of the new samples.

In this paper, we use the Gaussian Naive Bayes algorithm, which assumes that the values associated with each class are distributed according to a Gaussian distribution.

B. *Logistic Regression*

Logistic regression is a regression model that fits the values to the logistic function. It is useful when the dependent variable is categorical [5]. The general form of the model is

$$P(Y|X, W) = \frac{1}{1+e^{-(w_0+\sum w_i x_i)}} \quad (1)$$

Logistic regression is often used with regularization techniques to prevent overfitting. An L2 regularized model is used in this paper.

C. *Multi-layer Perceptron (MLP) Classifier*

MLP is an artificial neural network (ANN) model that consists of multiple layers of nodes, each fully connected to the next. The algorithm uses backpropagation for training the model[6]. The input is transformed using a learned non-linear transformation, which projects the input data into a space where it becomes linearly separable. This intermediate layer is called a hidden layer.

D. *K-Nearest Neighbors (KNN)*

The KNN algorithm classifies new data based on the class of the k nearest neighbors. This paper uses the value of k as 6. The distance from neighbors can be calculated using various distance metrics, such as Euclidean distance, Manhattan distance (used in this paper), Minkowski distance, etc. The class of the new data may be decided by majority vote or by an inverse proportion to the distance computed. KNN is a non-generalizing method, since the algorithm keeps all of its training data in memory, possibly transformed into a fast indexing structure such as a ball tree or a KD tree.

The Manhattan distance is computed using the formula

$$D = \sum |x_i - y_i| \quad (2)$$

IV. EXPERIMENTAL SETUP

The data was pulled from a sample dataset on Kaggle. There are two class labels—left and not left, labeled 1 and 0 respectively. The dataset had 14,999 data points, each labeled left or not left.

The dataset included various important features including average number of monthly hours, number of projects, years spent in the company and whether the employee received a promotion in the last five years. There were a total of nine features, out of which two were categorical and seven were numeric.

A. *Data pre-processing*

All the categorical values in each column were converted to numerical values by assigning integers to each category. For example, salary values, which were either 'low', 'medium', or 'high', were converted to 0, 1 and 2 respectively.

B. *Model Validation*

The dataset was split 70-30 into training and test sets. The models were trained using their optimal configurations on the training dataset. The trained model was then used to predict on the 30% test set.

The choice of model validation techniques in this paper is the area under the receiver operating characteristic curve (ROC-AUC). The AUC of a classifier is equal to the probability that the classifier will rank a randomly chosen positive example higher than a randomly chosen negative value [7].

Additionally, accuracy and F1 scores of the classifiers are also used to compare the results of the models. These two are important because they clearly show how suitable the model is for use in an application.

C. *System Environment Specification*

All classifiers are used from the scikit-learn package in Python 3.4. The code was run on an Debian GNU/Linux 8 system with 8 GB RAM.

V. RESULTS

TABLE I. MODEL RESULTS

| Algorithm | AUC | Accuracy | F1 Score |
|---|---|---|---|
| KNN | **0.9697** | **0.9432** | **0.8826** |
| Naive Bayes (Gaussian) | 0.8512 | 0.8029 | 0.6361 |
| Logstic Regression | 0.8078 | 0.7715 | 0.3482 |
| MLP Classifier (ANN) | 0.9176 | 0.8883 | 0.7834 |

A. *ROC Curves*

The ROC curve shows the general 'predictiveness' of the classifier. It measures the probability that the classifier ranks a randomly chosen positive instance higher than a randomly chosen negative instance. Closer the curve to the top left corner, better the classifier. The ROC curves for each classifier is shown in Figure 1.

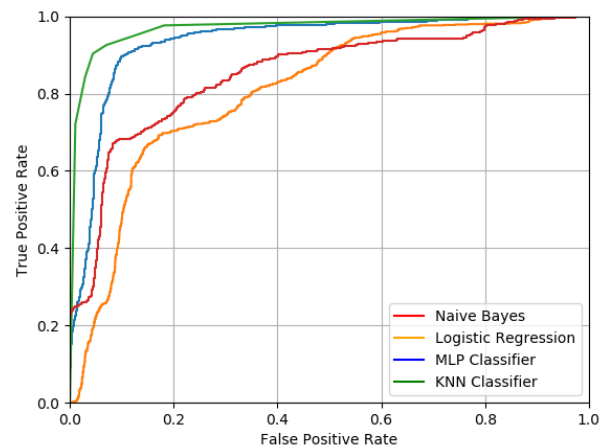

Fig. 1. ROC Curves for the Classifiers

B. *Discussion and Future Work*

The dataset is a good representative of the general workforce in today's organizations. The good results from multiple classifiers justify that the features chosen are causes that contribute to voluntary attrition.

Intuitively, data points that are close to each other are likely to have the same outcome of attrition. This is the basis for choosing the KNN algorithm in this paper. This intuition is validated by the observations of Figure 1, which are shown in Table 1. The KNN classifier has good ROC-AUC and accuracy values. Instead of constructing a general model, it simply stores instances of the data and classifies by a majority vote of the classes of the nearest neighbors.

Future work might include modifying the algorithm to weight neighbors so that nearer neighbors contribute more to the fit, rather than using uniform weights for all neighbors, and comparing results to the basic KNN model.

## VI. Conclusion

This paper presented the effect of voluntary attrition on organizations, and why predicting it is important. It further outlined various classification algorithms based on supervised learning to solve the prediction problem.

The results of this research showed the superiority of the KNN classifier in terms of accuracy and predictive effectiveness, by means of the ROC curve. When used with its optimal configuration, it is a robust method that delivers accurate results in spite of the noise in the dataset, which is a major challenge for machine learning algorithms. The authors thus recommend the use of the KNN classifier for accurately predicting employee attrition in an organization, which enables HR to take necessary action for the retention of employees predicted to be at risk of leaving.

## Acknowledgment

We thank our faculty, Ms. Saraswathi Punagin, for her guidance, encouragement, and co-operation throughout the completion of this paper. We also thank PES Institute of Technology for conducting RISE, which gave us an opportunity to present our work.

## References

[1] Cotton, J.L. and Tuttle, J.M., 1986. "Employee turnover: A meta-analysis and review with implications for research" Academy of management review, pp.55-70.

[2] Liu, D., Mitchell, T.R., Lee, T.W., Holtom, B.C. and Hinkin, T.R., 2012. "When employees are out of step with coworkers: How job satisfaction trajectory and dispersion influence individual-and unit-level voluntary turnover". Academy of Management Journal, pp.1360-1380.

[3] Heckert, T.M. and Farabee, A.M., 2006. "Turnover intentions of the faculty at a teaching-focused university". Psychological reports, pp.39-45.

[4] Rish, Irina, "An empirical study of the naive bayes classifier", IJCAI Workshop on Empirical Methods in AI.

[5] David A. Freedman, Statistical Models: Theory and Practice, Cambridge University Press p. 128.

[6] Rosenblatt, Frank, Principles of Neurodynamics: Perceptrons and the Theory of Brain Mechanisms.

[7] Fawcett, T., 2006. An introduction to ROC analysis. Pattern recognition letters, pp.861-874.